# Neutro-Connectedness Cut

Min Xian, Yingtao Zhang, H. D. Cheng, Fei Xu, Jianrui Ding

*Abstract*— **Interactive image segmentation is a challenging task and receives increasing attention recently; however, two major drawbacks exist in interactive segmentation approaches. First, the segmentation performance of ROI-based methods is sensitive to the initial ROI: different ROIs may produce results with great difference. Second, most seed-based methods need intense interactions, and are not applicable in many cases.**

**In this work, we generalize the Neutro-Connectedness (NC) to be independent of top-down priors of objects and to model image topology with indeterminacy measurement on image regions, propose a novel method for determining object and background regions, which is applied to exclude isolated background regions and enforce label consistency, and put forward a hybrid interactive segmentation method, Neutro-Connectedness Cut (NC-Cut), which can overcome the above two problems by utilizing both pixel-wise appearance information and region-based NC properties. We evaluate the proposed NC-Cut by employing two image datasets (265 images), and demonstrate that the proposed approach outperforms state-of-the-art interactive image segmentation methods (Grabcut, MILCut, One-Cut, MGC$_{max}^{sum}$ and pPBC).**

## I. INTRODUCTION

INTERACTIVE image segmentation, extracting the objects of interests from image with user interactions, is an essential and challenging task in image processing, and is also of great importance in object detection and recognition, biomedical image analysis, image and video editing, etc. Many inspiring interactive segmentation methods have been proposed, and can be classified into seed-based [1 - 5, 12, 13, 24 - 28] and region of interest (ROI)-based [6 - 8, 29, 32, 33] approaches in terms of the ways of interaction. Interactive image segmentation can achieve satisfactory results potentially because it can incorporate priori information (appearance, shape, topology, context, etc.) and correct errors by interactions. However, intense and precise user interactions are needed to segment complicated objects using seed-based approaches, which lead to low usability; although ROI-based methods need much less user interaction, their performances are sensitive to the initial ROI [34].

M. Xian, H. D. Cheng and F. Xu are with the Department of Computer Science, Utah State University, Logan, UT 84322 USA (e-mail: min.xian@ aggiemail.usu.edu; hd.cheng@ aggiemail.usu.edu; fei.xu@aggiemail.usu.edu).

Y. Zhang is with the School of Computer Science, Harbin Institute of Technology, Harbin, Heilongjiang 150001, China (e-mail: yingtao@hit.edu.cn).

J. Ding is with the School of Computer Science, Harbin Institute of Technology, Weihai, Shandong 264209, China (e-mail: jrding@hit.edu.cn).

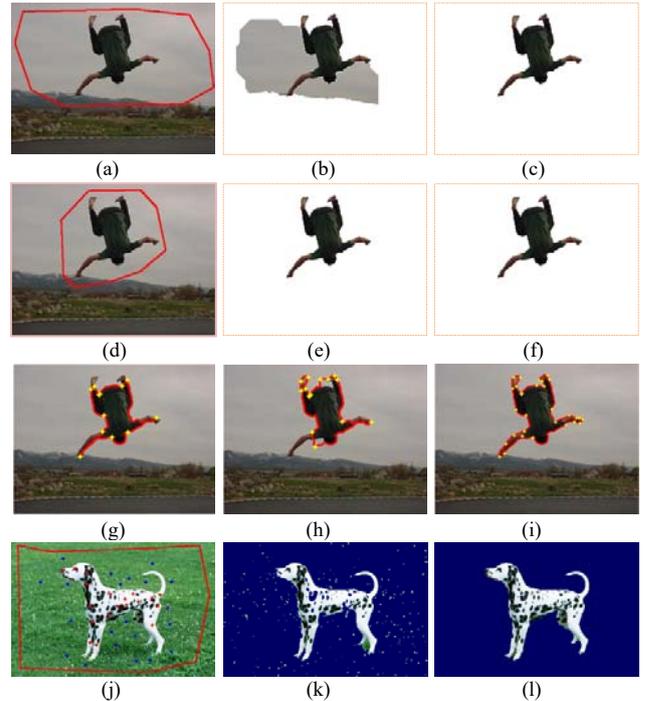

Fig. 1. The drawbacks of initial ROI-dependence and intense user interactions in interactive segmentation. (a), (d) and (j) two images with different user interactions; (b) and (e) the segmentation results of Grabcut [6]; (c), (f) and (l) the results of the proposed NC-Cut; (g), (h) and (i) the segmentation results (**red boundaries**) of intelligent scissors [13] with insufficient, incorrectly marked, and sufficient boundary seeds (**yellow points**), respectively; (k) the segmentation result of GC$_{max}^{sum}$ [5] using the marked seeds (red points in objects and blue points in background, respectively) in (j).

Seed-based methods can be classified into two subcategories: *boundary seed-based* methods characterized by specifying seeds on object boundary, and *region seed-based* methods needing user specified seeds in both object and background regions.

Live wire [12] and intelligent scissors [13] are two early boundary seed-based approaches. They imposed hard constraints on segmentation by marking certain seeds (anchor points) on object boundary. The final object boundary was obtained by computing the shortest path among these seeds. There are two limitations of these methods: first, insufficient boundary seeds may result in low segmentation accuracy (Fig. 1(g)); second, the seeds need to be marked on object boundary accurately (Fig. 1(h)), which is even impossible in many cases (low quality biomedical image segmentation, mobile device-based applications, etc.).

Graph cuts approach [1] is a *region seed-based* method to incorporate object and background priors with boundary con-straints into a combinational optimization framework in a soft way. The object seeds and background seeds will be utilized to estimate the object and background appearances (histogram, Gaussian Mixture Model (GMM)), respectively. In order to obtain accurate result, a large amount of seeds have to be marked; especially, when image has complicated texture structures in objects and/or background. Xiang et al. [24] formulate the interactive segmentation as a spline regression problem with nonlinear representation ability. The parameters in the spline function can be solved linearly by using the user specified object and background pixels, and the labels of all other pixels can be determined by the signs of spline function values easily. However, user had to specify quite many seeds to obtain controllable and desired accuracy.

In [25], Spina et al. proposed a hybrid interactive segmentation approach, Live markers, by combining the boundary seed-based methods (Live-wire-on-the-fly [26], Riverbed [27]) and region seed-based methods (Image Foresting Transform (IFT) [28], Graph cuts [1]). First, the method tracked object boundary by using user specified boundary seeds; then, the pixels adjacent to the tracked boundary were set as object or background seeds, which    initialize the IFT or Graph cuts methods. Although the method needs less user interaction than traditional boundary seed-based methods, it still has to specify seeds on the boundary accurately.

Grabcut [6] is one of the most popular ROI-based interactive segmentation methods.  The user-specified ROI is utilized for appearance model initialization: the pixels inside and outside the ROI are employed to estimate the object and background appearance models (GMMs), respectively. Then the method iterates graph cuts and appearance models re-estimation until convergence. The method needed much less user interaction and achieved better segmentation results than graph cuts. However, its performance is sensitive to the initial ROI: when the ROI does not cover the object, Grabcut will have high segmentation error rate.

In [29], Han et al. extended the color feature-based GMMs in Grabcut by constructing pixels' features based on color and multiscale nonlinear structure tensor [30].  The method can achieve better results than Grabcut for segmenting images with complex texture, but it needs a tight ROI. PinPoint [7] is another ROI-based method. It defined the tightness by employing a user-specified bounding box. The method assumes that the user specified ROI is tight enough; otherwise, it cannot obtain high segmentation accuracy.

MILCut [32] formulated the interactive segmentation as a multiple instance learning problem by generating negative and positive bags from the pixels outside a bounding box and the pixels of sweeping lines within the bounding box, respectively. The segmentation result is sensitive to the initial bounding box. One-Cut [8] was proposed to incorporate the measurement of L1 distance between object and background appearance models into the graph cuts energy function, and provided a graph construction method for high order potentials. It needs less user interaction than graph cuts. In [33], Tang et al. proposed a parametric Pseudo-Bound Cuts (pPBC) method for optimizing interactive segmentation with high-order and non-submodular energies. The experiments in section IV demonstrate that pPBC achieves better results than other ROI-based methods (Grabcut, MILCut and One-Cut); however, its performance is still sensitive to the initial ROI.

Deformable model-based methods [9-11] utilize the boundary of user-specified ROI as the initial contour. An initial contour close to the object boundary is necessary to make the segmentation converge to the real object boundary.

As mentioned above, intense user interactions are required to segment complicated objects using region seed-based approaches and precise interactions are needed in boundary seed-based methods, which lead to low usability of these methods; ROI-based methods need much less user interaction, but their performances are  sensitive to users' inputs. Fig. 1 illustrates the problems of interactive image segmentation by using three interactive methods: Grabcut [6], intelligent scissors [13], and $GC_{max}^{sum}$ [5]. As shown in Figs. 1(a), (b), (d) and (e), when the ROI covers the object loosely, the Grabcut's result is not good (Fig. 1(b)) which contains many background pixels; however, when the ROI becomes tight, the result is quite accurate (Fig. 1(e)). In Fig. 1(g), intelligent scissors tends to generate short boundary when insufficient seeds are specified, which misclassified some object regions; and if some boundary seeds are not on object boundary precisely, the segmentation result of intelligent scissors is not accurate (Fig. 1(h)). Fig. 1(i) shows that intelligent scissors can produce accurate results with 37 precisely specified seeds on the object boundary. Figs. 1(j) - (l) compares $GC_{max}^{sum}$ [5] with the proposed NC-Cut.

In this paper, we propose a novel hybrid interactive image segmentation approach, Neutro-Connectedness Cut (NC-Cut), which formulates segmentation based on both pixel-wise appearance models and NC properties of the regions. The NC is a global topologic property among image regions and can reduce the dependence of segmentation performance on the appearance models generated by user interactions. The user interaction is to specify a polygon containing the object, and the image regions outside the polygon are viewed as the background seeds. The proposed region-based NC computation algorithm calculates the NC between each region in the polygon and background seeds, and generates a NC forest (Fig. 4(d) and Figs. 5(b) and (d)) rooted from the background seed set. The constructed graph based on NC forest imposes NC as a global constraint to the segmentation. Moreover, we propose a novel method for enforcing label consistency of object regions and excluding isolated background region by using the pruning and linking operations (section III(C)). The proposed optimization algorithm estimates pixel labels, appearance models, NC, and object and background regions jointly.

The rest of the paper is organized as follows: the related work is discussed in section II; the proposed method is presented in section III; and the experimental results and conclusions are discussed in sections IV and V, respectively.

## II. RELATED WORK

*Connectedness* is an important global topology property and has a distinctive characteristic independent of the feature distributions in image appearance models, e.g., histogram, GMM, etc. Thus it has been widely employed in image segmentation [5, 14-16].

In classic logic, the connectedness between any pair of elements is defined as whether there exists a path between them. In fuzzy logic, fuzzy connectedness (FC) is defined as the degree of connectedness between two elements. The FC on color image was first proposed in [14].

In [5], an interactive segmentation method based on FC [14, 15] and Graph cuts [1] was proposed. First, object seeds and background seeds are manually selected for computing the object and background connectednesses, respectively. Then, some object pixels and background pixels will be chosen to estimate the object and background according to the relative FC values. The method can reduce user interactions; especially, when objects and background are relatively homogeneous, and can achieve better results than that of graph cuts. However, the intense user interactions are still needed; especially, for some natural image segmentation. Even many object seeds and background seeds are employed as shown in Fig. 1(j), the result still has too much noise as shown in Fig. 1(k) since it failed to handle the indeterminacy of connectedness.

In [16], an early definition of NC was proposed to generalize the FC by introducing *the indeterminacy domain* for measuring the uncertainty of the connectedness.

Protiere et al. proposed an interactive segmentation method based on geodesic distance [31]. In order to segment textures, the weights between adjacent pixels were calculated by using Garbor-based features rather than using the gradient.

In digital topology, *connectivity* defines the condition of adjacency between points, which is similar to the connectedness defined in classic logic. *Connectivity* [2, 3, 35, 36] is usually employed to solve the "shrinking bias" problem in Graph cuts. The method in [2], imposed connectivity constraint by user specified nodes connected to the foreground; however, setting extra seeds at the end of every elongated part of an object is a tough job. In [3], the connectivity constraint was formulated using a shortest geodesic path tree computed by Dijkstra's algorithm, and user only utilized a point or a region as the root node. The method cannot achieve good performance for segmenting objects with complicated appearance. Topology constraints based on connectivity and simple point have been considered in the max flow algorithm [35], and only local minimum can be obtained. Connectivity regularizer was utilized to build potential function to output connected labels [36]. In [37], the authors extended the star-convexity shape prior from single center to multiple centers and generalized the Euclidean-based star [38] to geodesic-based, which can reduce interaction effort largely. The geodesic trees in [37] are constructed based on geodesic distance which has problems discussed in the next paragraph.

The geodesic distance accumulates the distances between adjacent nodes along the shortest path [2, 3]. On a path, the farther a node from the source, the larger distance the node has. The geodesic distance can accumulate small errors and has a bias that favors the shorter paths [39], which makes geodesic distance-based segmentation approaches sensitive to seed placement. The connectedness is a global topological property; and the main differences between the connectedness and the geodesic distance are that the connectedness does not accu-mulate neighboring differences and is independent of the length of the path. The FC calculates the degree of connectedness of a path by using the two adjacent nodes with the least similarity, which makes the FC-based segmentation methods suffer from leaking [5, 16] and noise (Figs. 1(j) - (k)). The NC generalizes the FC, and introduces the degree of indeterminacy ($I$) to represent the uncertainty of the connectedness. In the proposed NC-Cut, we will use $I$ to adjust the weights of the connectedness term (Eqs. (14) - (15)). The NC-Cut will transfer more control to the appearance models in inhomogeneous regions (high $I$ values). Therefore, the proposed NC-Cut can handle both homogeneous and inhomogeneous images well. A shortest path with four nodes is shown in Fig. 2, and the distance between each pair of adjacent nodes ($d_{ij}$) is 0.1; and for simplicity, the degree of connectedness between adjacent nodes is defined by 1 - $d_{ij}$ and $I_n$ denotes the degree of indeterminacy of connectedness which will be described in section III(A). In Fig. 2, node 3 is affected by noise; the geodesic distance accumulates the error to node 3 and node 4; and FC cannot reflect the effect of the noise. The proposed NC introduces $I$ to represent the indeterminacy of the connectedness for handling the noise.

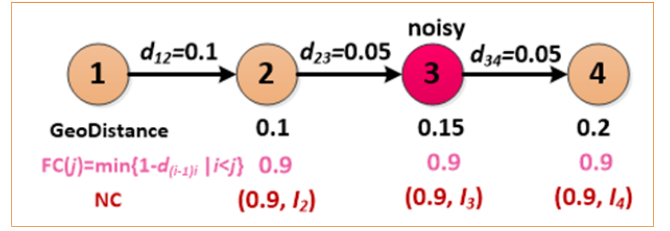

Fig. 2. Difference among Geodesic distance, fuzzy connectedness, and Neutro-Connectedness.

## III. PROPOSED METHOD

The major cause of the drawbacks of interactive segmentation methods is their heavy dependence on the object and background appearance models estimated directly from user interactions. When user-specified seeds are insufficient or the ROI does not enclose the object tightly, these approaches will have poor performance.

In this section, we present the newly proposed NC-Cut method for interactive image segmentation. The NC expands outward from the background regions (seeds) to generate the NC maps and NC forest representing the topological property of the image. As discussed in section II, NC does not have the shorter path bias as the geodesic distance, and can handle uncertainty better by introducing the indeterminacy. Therefore, NC is more suitable than geodesic distance and FC for interactive image segmentation. In addition, the NC is not defined on the appearance models initialized by user interaction and produces informative NC maps and NC forest requiring much less user interaction (Fig. 5). Therefore, by incorporating NC and image appearance models, the proposed NC-Cut requires much less user interaction than the seed-based methods, and is less sensitive to initial ROI than the existing ROI-based methods.

The flowchart of the proposed NC-Cut method is shown in Fig. 3.

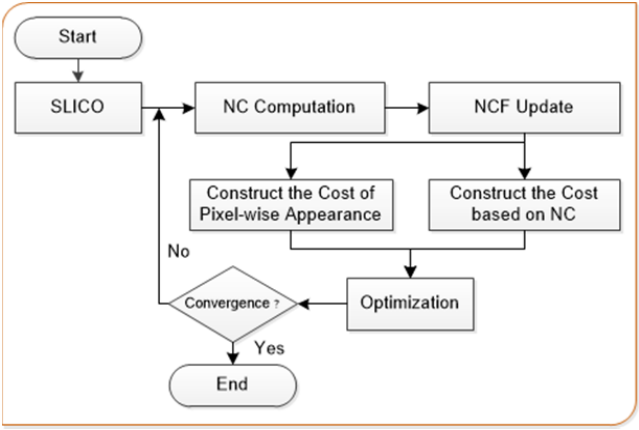

Fig. 3. Flowchart of the proposed method.

## A. Generalized Neutro-Connectedness

Neutro-connectedness (NC) was first proposed in [16], and applied to solve the weak boundary problem in breast ultrasound (BUS) image segmentation. The NC in [16] is a top-down method based on the priori information of the object and background. In this section, we will generalize the NC definition in [16] to make it independent of the top-down priors of objects.

We define the generalized NC on image regions produced by using the zero parameter version of SLIC (SLICO) [17].

**Definition 1:** Let $\mho$ be the set of all regions in an image, and $Z_i$ be the $i$th path between two regions in $\mho$. The strength of the *Neutro-Connectedness* of $Z_i$ is a triple defined by

$$NC_{path}(Z_i) = \left(\Pi^T(Z_i), \Pi^I(Z_i), \Pi^F(Z_i)\right)$$
$$Z_i = (z_1, z_2, \cdots, z_L), z_k \in \mho, k = 1, 2, \cdots, L \quad (1)$$

where L is the number of regions on $Z_i$, and $z_k$ denotes the $k$th region, $\Pi^T(Z_i)$, $\Pi^I(Z_i)$ and $\Pi^F(Z_i)$ represent the degree of the truth, indeterminacy and falsity of the connectedness of path $Z_i$, respectively. They are defined as follows.

$$\Pi^T(Z_i) = min\{\mu_T(z_j, z_k) | z_k \in \mathbb{N}(z_j), z_j, z_k \in Z_i\}$$
$$\Pi^I(Z_i) = max\{\mu_I(z_j, z_k) | z_k \in \mathbb{N}(z_j), z_j, z_k \in Z_i\} \quad (2)$$
$$\Pi^F(Z_i) = 1 - \Pi^T(Z_i)$$

In Eq. (2), we use $\mathbb{N}(z_j)$ to represent the set of all neighbors of region $z_j$, $\mu_T$ to represent the strength of connectedness between adjacent regions, and $\mu_I$ to represent the degree of indeterminacy of the connectedness. $\mu_T$ and $\mu_I$ are defined by

$$\mu_T(p, q) = e^{-\|m(p) - m(q)\|^2 / 2\delta_t^2} \quad (3)$$
$$\mu_I(p, q) = max\{h(p), h(q)\} \quad (4)$$

where $m(p)$ and $m(q)$ are the mean values of color features (RGB) of regions $p$ and $q$, respectively; $h(p)$ is the degree of inhomogeneity of region $p$. Eq. (3) penalizes the color discontinuity between regions $p$ and $q$: if $/m(p) - m(q)/ < \delta_t$, Eq. (3) penalizes more; otherwise less. $\mu_I$ is defined on the inhomogeneity of two adjacent regions, and the regional inhomogeneity is defined by [23]

$$h(p) = \frac{1}{n_1} \sum_{i=1}^{n_1} (f_{std}(i) \times f_{sobel}(i)) \quad (5)$$

where $n_1$ is the number of the pixels of region $p$, and $f_{std}(i)$ and $f_{sobel}(i)$ are the local standard deviation and sobel filter output at point $i$, respectively. In [16], $\mu_T$ and $\mu_I$ are defined on tissue appearance, which makes the method only segment a certain type of objects (i.e., breast tumors). However, the newly proposed NC is independent of object appearance.

**Definition 2:** Let $\mathbb{Z}$ be the set of all the paths between regions $p$ and $q$ in set $\mho$. The *Neutro-Connectedness* between $p$ and $q$, $NC(p, q)$, is defined in Eq. (6).

$$NC(p, q) = ((T, I, F) = NC_{path}(Z_d)), Z_d \in \mathbb{Z},$$
$$\forall Z_i \in \mathbb{Z}, \langle \Pi^T(Z_i), 1 - \Pi^I(Z_i) \rangle \preccurlyeq \langle \Pi^T(Z_d), 1 - \Pi^I(Z_d) \rangle \quad (6)$$

In Eq. (6), $Z_d$ is the path with the strongest connectedness between regions $p$ and $q$ in set $\mathbb{Z}$; $T$, $I$ and $F$ are the corresponding degrees of the truth, indeterminacy and falsity of the connectedness between regions $p$ and $q$, respectively; the operator $\preccurlyeq$ denotes a lexicographical order relation defined as

$$\langle a_1, b_1 \rangle \preccurlyeq \langle a_2, b_2 \rangle \iff a_1 < a_2 \text{ or } (a_1 = a_2 \text{ and } b_1 \leq b_2).$$

---

**Algorithm 1: Region-based Neutro-Connectedness**

Inputs: $AT = [\mu_T(p, q)]_{N \times N}$, $AI = [\mu_I(p, q)]_{N \times N}$,
$SRs = \{sr_k\}_{k=1}^n$
Outputs: $\{(T_r, I_r)\}_{r=1}^N$, $NCF = \{(pre_r, rt_r)\}_{r=1}^N$

  Initialization: $(T_r, I_r) = (0, 0), r = 1, 2, \cdots, N$
    $\forall sr_k \in SRs, (T_{sr_k}, I_{sr_k}) = (1, AI(sr_k, sr_k))$,
    $pre_{sr_k} = sr_k, rt_{sr_k} = sr_k,$
    Put all regions in SRs to queue $Q$.
NC computation:
1: Extract the region $p$ with the strongest connectedness
  on $Q$ according to $\preccurlyeq$
2: For each region $q$ adjacent to $p$
    if $\langle T_q, 1 - I_q \rangle \prec$
    $\langle min\left(T_p, AT(p, q)\right), 1 - max\left(I_p, AI(p, q)\right) \rangle$ then
      $T_q = min\left(T_p, AT(p, q)\right)$
      $I_q = max\left(I_p, AI(p, q)\right), pre_q = p, and rt_q = rt_p$
      Insert $q$ to $Q$.
3: Repeat steps 1 and 2 until $Q$ is empty.

---

## B. Neutro-Connectedness Computation

We propose a region-based NC computation algorithm (Algorithm 1) which outputs NC quite fast and costs much less space than the method in [15]; moreover, Algorithm 1 generates the NC forest to explore the topological structure of NC.

As shown in Definitions 1 and 2, the NC between two regions is defined as a triple $(T, I, F)$, and $F$ is defined as $1 - T$; therefore, we only need to calculate $(T, I)$ for further discussion. Algorithm 1 computes the NC between each region and the seed regions, AT and AI are two sparse matrices; AT saves the dissimilarities between every pair of adjacent regions calculated by using Eq. (3); AI saves the degrees of indeterminacy (Eq. (4) and Eq. (5)) of dissimilarities between adjacent regions; SRs is a set of seed regions generated automatically or specified by users; $(T_r, I_r)$ is the NC between the $r$th region and the SRs; and $pre_r$ and $rt_r$ denote the parent and root nodes of node $r$, respectively. The operator $\prec$ returns true when $\langle a_1, b_1 \rangle \preccurlyeq \langle a_2, b_2 \rangle$ and $b_1 \neq b_2$.

Algorithm 1 is a modified version of Dijkstra's method and the best time complexity can be linear [40]; however, in FC, $N$

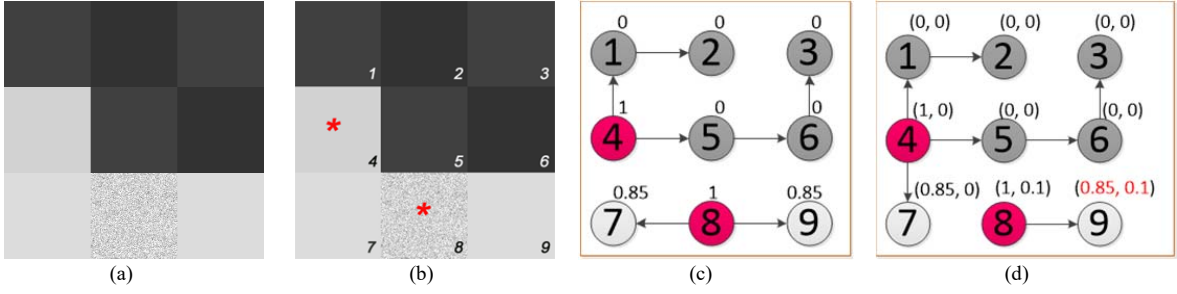

Fig. 4. Neutro-Connectedness computation of a synthetic image. (a) A synthetic image with nine regions; (b) the region index (lower right corner) and seed regions (indicated by red '*'); (c) the forest and $T$ map (FC) generated without considering the indeterminacy ($I$); (d) Neutro-Connectedness forest (NCF) and ($T$, $I$) map produced by using Algorithm 1.

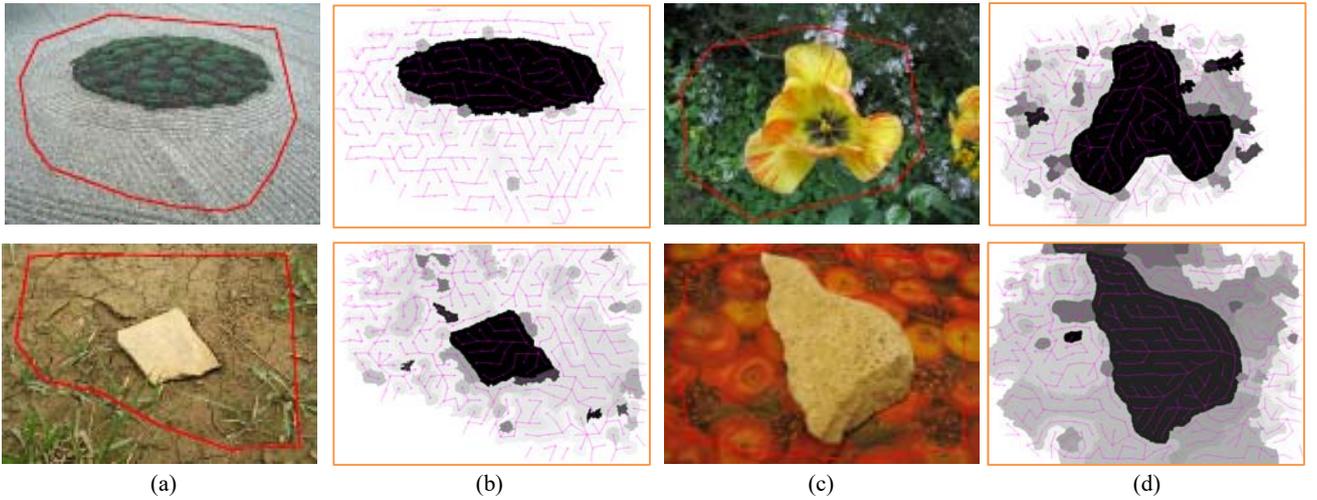

Fig. 5. NC computation of natural images. (a) and (c) four original images with user specified ROIs (red polygons); (b) and (d) NCFs of the maps of the degree of truth of NC for each region ($T_i$ map).

is the number of pixels (e.g. 500×500) which is usually at least two orders of the magnitude greater than the number of image regions (500 in the proposed method). Therefore, the proposed method is faster and costs much less memory space than FC. The image foresting transform (IFT) [28] generates spanning forest for image, which is similar to NCF generation process in Algorithm 1, and a superpixel implementation of the IFT can be found in [40]; nevertheless, there are two main differences between NCF and IFT: first, the fundamental concept for constructing NCF is taking both the truth and indeterminacy of the connectedness into consideration, while IFT constructs spanning forests by selecting the path with the minimum path cost. Second, in the case of ties (two paths with same $T$ value), IFT follows the last-in-first-out (LIFO) queue policy to break ties; Algorithm 1 introduces the lexicographical order relation of ($T$, $I$) to determine the best path.

Fig. 4 shows an example of NC computation of a synthetic image with nine regions falling into two categories: dark regions (1, 2, 3, 5, and 6) and light regions (4, 7, 8, and 9). Note that region 8 is an inhomogeneous region with Gaussian noise. The fourth and eighth regions are the seed regions ($SRs$ = {4, 8}). Fig. 4(c) shows the forest (FC-based) and the degree of connectedness ($T$) between each region and the seed regions without considering the indeterminacy ($I$), which is the same with the results of FC algorithm; all the light regions have high

$T_i$ values, and all the dark regions have low $T_i$ values; regions 7 and 9 are both connected to seed region 8, and have the same degree of connectedness. Fig. 4(d) shows results of the proposed NC; the NCF in Fig. 4(d) is different from the forest in Fig.4(c); even though regions 7 and 9 have same $T$ values (0.85) to the seed regions, the degree of indeterminacy of region 9 (0.1) is higher than that of region 7 (0) due to the noise.

Fig. 5 shows the results of NC computation of the natural images. All the regions outside ROI are the background seeds in $SRs$; the forests, NCFs, are shown in the corresponding $T_i$ maps. As shown in Figs. 5(b) and (d), all trees are rooted from $SRs$; the $T_i$ value of a node is less than or equal to that of its parent node; the object regions have low $T_i$ values; however, some background regions may also have low $T_i$ values if they are isolated. Such problem will be solved in section III(C).

Similar to [15], we design Algorithm 1 using the dynamic programing strategy. There are three main differences between the proposed NC algorithm and the FC algorithm: *first*, the proposed algorithm generates a region-based NC forest, NCF, which uncovers the topological structure of NC maps by using trees; *second*, the proposed algorithm calculates the degrees of the truth, indeterminacy and falsity of connectedness simultaneously; *third*, the FC method calculates the connectedness between pixels; while the proposed method calculates the connectedness between regions which reduces the computation time and memory cost.

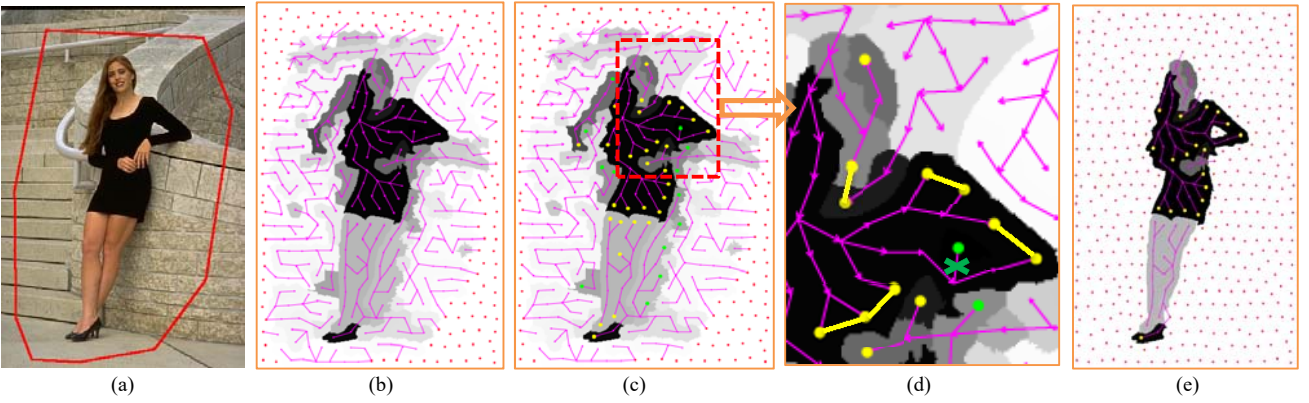

Fig. 6. Candidate object and background regions. (a) Original image with a loose ROI (red polygon); (b) background NCF (magenta arrows); (c) the object (yellow points) and background (green points) regions on $T_i$ map; (d) the pruning (green cross) and linking (yellow edges) operations of the region in the red dashed rectangle in (c); (e) the candidate object and background regions after the fifth iteration of Algorithm 2 (III(F)).

### C. NCF Update

The background Neutro-Connectedness between each region in ROI and the background seed regions ($SRs^{bkg}$), is computed by setting the regions outside the specified ROI as the initial $SRs^{bkg}$ and invoking Algorithm 1. Algorithm 1 also outputs a background Neutro-Connectedness forest (NCF$^{bkg}$), in which the root of every tree is from the background $SRs$. As shown in Figs. 5(b) and (d), all object regions have quite low connectedness to $SRs^{bkg}$, and are on the same tree.

However, as shown in Figs. 5(b) and (d) and 6(b), some background regions may have low connectedness if they are not connected to the initial background $SRs^{bkg}$. The segmentation method only using the background NC cannot handle these isolated regions. In order to cope with such problem, we update NCF by integrating pruning and linking operations.

#### 1) Object and Background Regions

In every tree of the NCF$^{bkg}$, the leaf nodes (regions) have the lowest connectedness ($T$ value) to their roots (background seeds), and have the highest possibility to be object or isolated background regions; consequently, the candidate object and background regions are defined by

$$P^{obj} = \{r \mid r \text{ is a leaf region in NCF}^{bkg} \text{ and } \\ T_r^{bkg} < \tfrac{1}{N}\sum_{i=1}^{N} T_i^{bkg}, \text{ and } r \notin B\} \quad (7)$$

and

$$P^{bkg} = \{r \mid r \text{ is a leaf region in NCF}^{bkg} \text{ and } \\ T_r^{bkg} < \tfrac{1}{N}\sum_{i=1}^{N} T_i^{bkg} \text{ and } r \in B\} \quad (8)$$

respectively, where $T_r^{bkg}$ is the degree of the truth of connectedness between the $r$th region and the background seed regions, the threshold for filtering the leaf regions is set as the mean of all regions' background connectedness ($\tfrac{1}{N}\sum_{i=1}^{N} T_i^{bkg}$), and set $B$ includes all regions of high similarity with background seed regions and is given by

$$B = \{r \mid f(r) = e^{-\left(p^{bkg}(m(r)) - u_B\right)^2 / (2\delta_B^2)} > \varepsilon\} \quad (9)$$

where $p^{bkg}(\cdot)$ is the background GMM distribution learned from $SRs^{bkg}$, $m(r)$ is the mean of color features (RGB) of the $r$th region, $u_B$ is the mean of $p^{bkg}(\cdot)$ of the regions in $SRs^{bkg}$ and is defined as $(1/n^{bkg})\sum_{r \in SRs^{bkg}} p^{bkg}(m(r))$ where $n^{bkg}$ is the size of $SRs^{bkg}$, and $\delta_B$ and $\varepsilon$ are the standard deviation and the threshold, respectively.

The candidate object regions $P^{obj}$ include all the leaf regions in NCF$^{bkg}$ having low $T$ value and low similarity with $SRs^{bkg}$, and the candidate isolated background regions $P^{bkg}$ are the leaf regions with low connectedness but high similarity with $SRs^{bkg}$.

As shown in Fig. 6 (c), 29 of 30 candidate regions of the object (indicated by **yellow points**) determined by Eq. (7) belong to the object; and all candidate regions of isolated background (indicated by **green points**) determined by Eq. (8) are in the background. In Fig. 6(e), after five iterations of Algorithm 2 (III(F)), all candidate object regions are correctly in the object; and there is no candidate background regions because no region satisfies Eq. (8).

#### 2) NCF update

Two operations exist in the NCF update process:

1) Pruning. Disconnect the edge between any isolated background region and its parent node by resetting its parent and root nodes to itself (Eqs. (10) and (11)).

$$npre_r = \begin{cases} r, & r \in P^{bkg} \\ pre_r, & otherwise \end{cases} \quad (10)$$

$$nrt_r = \begin{cases} r, & r \in P^{bkg} \\ rt_r, & otherwise \end{cases} \quad (11)$$

In Eqs. (10) and (11), $npre_r$ and $nrt_r$ denote the new parent node and the new root node of the $r$th region.

2) Linking. Add an *auxiliary edge* to each pair of adjacent object regions located on the same tree (Eq. (12)).

$$aux_r = \begin{cases} g, & g \in \mathbb{N}(r)r, g \in P^{obj}, and\ rt_r = rt_g \\ r, & otherwise \end{cases} \quad (12)$$

In Eq. (12), $\mathbb{N}(r)$ denotes all the adjacent regions of the $r$th region.

The pruning operation is applied for breaking the link between an isolated background region and its parent, which will exclude the isolated background regions from the segmentation results. The linking operation connects object nodes by adding *auxiliary edges*, which will enforce label consistency to adjacent object nodes. The modified NCF (nNCF) is defined by

$$nNCF = \{(npre_r, nrt_r, aux_r)\}_{r=1}^{N} \quad (13)$$

We use Figs. 6(c) and (d) to demonstrate how to update the original NCF. In Fig. 6(d), the yellow line segments indicate the auxiliary edges added by the linking operation, and the green cross mark indicates that the edge will be removed by the pruning operation.

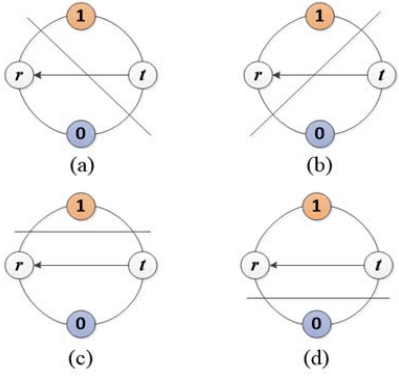

Fig. 7. Four possible labeling schemes (cuts) for two adjacent nodes in NCF. (a) node $r$ is labeled as $\mathbf{0}$ and $t$ as $\mathbf{1}$; (b) node $r$ is labeled as $\mathbf{1}$ and $t$ as $\mathbf{0}$; (c) both $r$ and $t$ are labeled as $\mathbf{0}$; and (d) both $r$ and $t$ are labeled as $\mathbf{1}$.

### D. Neutro-Connectedness Cut Formulation

Let $S^p = \{s_i^p \in \{\mathbf{0},\mathbf{1}\}\}_{i=1}^{N}$ be a group of binary labels for the pixels: labels $\mathbf{0}$ and $\mathbf{1}$ are for pixels in background and object, respectively; Graph $G_{nNCF}$ is built on the modified NC forest, and $G_A$ is constructed as the same as that in [1]. The Neutro-Connectedness Cut (NC-Cut) is formulized as

$$min\, E(S^p, \theta^A, w) = E^{G_A}(S^p, \theta^A) + \gamma E^{G_{nNCF}}(S^p, w^p) \quad (14)$$

In Eq. (14), the cost function $E$ is defined based on both appearance and topological properties; $E^{G_A}$ takes the form of Grabcut's cost function [6] and defines the cost according to the region and boundary properties; $E^{G_{nNCF}}$ defines the cost of a cut on $G_{nNCF}$; $w^p$ defines the weights of t-links (terminal links) and n-links (neighborhood links) of $G_{nNCF}$; $\gamma$ controls the contribution of $E^{G_{nNCF}}$, and is defined by

$$\gamma = e^{-(\bar{I})^2/(2\delta_\gamma^2)} \quad (15)$$

where $\bar{I}$ is the mean value of the indeterminacies of all regions; $\delta_\gamma$ is the standard deviation; and $\theta^A$ is the set of object and background GMM parameters defined by

$$\theta^A = \{(\pi(k,s), \mu(k,s), \Sigma(k,s)) | k = 1, \cdots, K, s \in \{0,1\}\} (16)$$

In Eq. (16), K is the number of components of object (or background) GMM; $\pi(k,s), \mu(k,s),$ and $\Sigma(k,s)$ are the Gaussian mixture weight, mean and covariance matrices of the $k$th GMM component for object ($s = 1$) or background ($s = 0$), respectively.

### E. $G_{nNCF}$ Construction

Let $G_{nNCF} = (\mathcal{V}, \mathcal{E})$ be a graph with vertex set $\mathcal{V}$ (regions) and edge set $\mathcal{E}$ defined on the nNCF. The edge set $\mathcal{E}$ is defined as

$$\mathcal{E} = \underbrace{\{(r,t) | r \neq t \text{ and } (npre_r = t \text{ or } aux_r = t)\}}_{\mathcal{E}_n:\, n\text{-links}} \cup$$
$$\underbrace{\{(r,0)\} \cup \{(r,1)\}}_{\mathcal{E}_t:\, t\text{-links}}, r = 1,2,\cdots, N \quad (17)$$

where $npre_r$ is the parent node of the $r$th region, and $aux_r$ is the region having an auxiliary edge with the $r$th region in nNCF.

Let $w(r,t)$ be the nonnegative weight for each edge $(r,t) \in \mathcal{E}$ of graph $G_{nNCF}$. A cut $C$ is a subset of $\mathcal{E}$, which partitions the vertex set into two disjoint parts. The cost of cut $C$ is the sum of the weights of the edges in $C$, and the optimal cut will have the minimal cost.

We introduce the local connectedness constraint (**C1**) to the cut $C$: let $(r,t) \in \mathcal{E}$ and $t = npre_r$, if node $r$ is labeled as 0 (background), the label of node $r$'s parent node $t$ should also be 0. If node $t$ is the parent node of node $r$, there will be four possible labeling schemes (Fig. 7). The costs of the four possible cuts are

$$|C_a| = w(r,\mathbf{1}) + w(t,\mathbf{0}) + w(r,t),$$
$$|C_b| = w(r,\mathbf{0}) + w(t,\mathbf{1}) + w(r,t),$$
$$|C_c| = w(r,\mathbf{1}) + w(t,\mathbf{1}), \text{and} \quad (18)$$
$$|C_d| = w(r,\mathbf{0}) + w(t,\mathbf{0}).$$

As shown in Fig. 7, the cuts in Figs. 7(b) - (d) satisfy **C1**, but Fig. 7(a) does not. In order to avoid the cut in Fig. 7(a), the optimal cut $C_\kappa$ must satisfy

$$|C_\kappa| = w(r,s_r) + w(t,s_t) + |s_r - s_t| w(r,t)$$
$$< w(r,\mathbf{1}) + w(t,\mathbf{0}) + w(r,t) = |C_a| \quad (19)$$

The weights of t-links in $G_{nNCF}$ are defined by

$$w(r,s_r) = \begin{cases} -log\, T_r^{bkg}, & if\ s_r = \mathbf{1} \\ -log\left(1 - T_r^{bkg}\right), & if\ s_r = \mathbf{0} \end{cases}, r = 1,2,\cdots, N \ (20)$$

where $T_r^{bkg}$ is the degree of the truth of connectedness between the $r$th region and the seed regions in background; the weights of n-links are defined by

$$w(r,t) = \begin{cases} \lambda e^{-(T_r^{bkg} - T_t^{bkg})^2/(2\delta_{NC}^2)}, & if\ t = npre_r \\ \lambda, & if\ t = aux_r \\ 0, & otherwise \end{cases} \quad (21)$$

where $\delta_{NC}$ is set as 0.1 by experiment, and $\lambda$ is a fixed value larger than max $\{w(r,s_r) | r = 1,2,\cdots, N\}$.

**Theorem 1.** Let $G_{nNCF} = (\mathcal{V}, \mathcal{E})$ be a graph constructed according to the background Neutro-Connectedness. If the edge weights $w$ are defined by Eqs. (20) and (21), then there always exists a local cut satisfying **C1**.

**Proof.** Let $t$ be the parent node of $r$ in a background NCF. Then $1 \geq T_t \geq T_r \geq 0$.

If $T_t > T_r$, we get $w(t,\mathbf{1}) > w(t,\mathbf{1})$ and $w(t,\mathbf{0}) > w(r,\mathbf{0})$, therefore $|C_b| < |C_a|$;

If $T_t = T_r$, we have

$w(r,\mathbf{1}) = w(t,\mathbf{1}), w(r,\mathbf{0}) = w(t,\mathbf{0})$ and $w(r,t) = \lambda$, therefore $|C_c| = w(r,\mathbf{1}) + w(t,\mathbf{1}) = 2w(r,\mathbf{1})$
$$< w(r,\mathbf{1}) + w(r,\mathbf{0}) + \lambda = |C_a| \text{ and}$$
$$|C_d| = w(r,\mathbf{0}) + w(t,\mathbf{0}) = 2w(r,\mathbf{0})$$
$$< w(r,\mathbf{1}) + w(r,\mathbf{0}) + \lambda = |C_a|$$

Hence, there always exits a local cut whose cost is smaller than $|C_a|$, i.e., **C1** is satisfied.

If $(r,t) \in \mathcal{E}$ and $t = aux_r$, it means that $r$ and $t$ are two object regions and $(r,t)$ is an auxiliary edge produced by the linking operation, edge $(r,t)$ has high weight ($w(r,t) = \lambda$); therefore, similar to the second situation ($T_t = T_r$) in the above proof, a local cut tends to assign label $\mathbf{1}$ to both $r$ and $t$, which enforces label consistency to object regions with auxiliary edge.

The computation of NC and NCF, and construction of $G_{nNCF}$ are based on image regions to prevent the high time and memory cost of pixel-wise NC computation. However, as shown in Eq. (14), the cost function of the NC-Cut is formulated on image pixels to avoid assigning a uniform label to all pixels in an inhomogeneous region. Therefore, we approximate the NC of pixel by Eqs. (22) and (23). The pixel-wise cost function $E^{G_{nNCF}}$ of $G_{nNCF}$ is defined by

$$E^{G_{nNCF}}(S^p, w^p) =$$
$$(\sum_{i=1}^{NP} w^p(i, s_i^p) + \eta \sum_{i=1}^{NP} \sum_{j \in N_i} |s_i^p - s_i^p| w^p(i, j)) \quad (22)$$

In Eq. (22), $w^p(i, s_i^p)$ defines the weight of the t-link between the $i$th pixel and $s_i^p$ (0 or 1); and $w^p(i, j)$ defines the weight of n-link between the $i$th and $j$th pixels. If the $i$th pixel is in the $r$th region, then $w^p(i, s)$ is given by

$$w^p(i, s_i^p) = w(r, s_i^p), \quad (23)$$

where $w(r, s_i^p)$ is defined in Eq. (20); and $w^p(i, j)$ is defined by

$$w^p(i, j) = \begin{cases} \lambda, & if\ R(i) = R(j) \\ w(R(i), R(j)), & j \in N_i\ and\ (R(i), R(j)) \in \mathcal{E}_n \\ 0, & otherwise \end{cases} \quad (24)$$

where $N_i$ is the set of $i$'s adjacent pixels, and $R(i)$ is the index of the region including the $i$th pixel.

$E^{G_A}$ is defined as [6]

$$E^{G_A}(S^p, \theta^A) = \sum_{i=1}^{NP} -log(p_G(x_i, \theta^A, s_i^p))$$
$$+ \eta \sum_{i=1}^{NP} \sum_{j \in N_i} |s_i^p - s_i^p| e^{\beta \|x_i - x_j\|^2} / D(i, j) \quad (25)$$

where $x_i$ is the color feature (RGB) of the $i$th pixel, $\eta$ is set as 50 [6]; the constant $\beta$ is set in the same way as in [6]; $D(i, j)$ is the Euclidean distance between the $i$th and the $j$th pixels; and $p_G$ outputs the possibility of the $i$th pixel belonging to the object or background, and is defined by [6]

$$p_G(x_i, \theta^A, s_i^p) =$$
$$\max \{ \pi(k, s_i^p) \mathcal{N}(\mu(k, s_i^p), \Sigma(k, s_i^p)) \}, k = 1, 2, \cdots K \quad (26)$$

where $\mathcal{N}(\mu(\cdot), \Sigma(\cdot))$ is a multi-dimensional Gaussian distribution with mean vector $\mu$ and covariance matrices $\Sigma$; $\pi(k, s_i^p), \mu(k, s_i^p)$, and $\Sigma(k, s_i^p)$ are the Gaussian mixture weight, mean and covariance matrices of the $k$th GMM component for object ($s_i^p = 1$) pixel or background ($s_i^p = 0$) pixel, respectively.

The cost function in Eq. (14) can be rewritten as

$$E(S^p, \theta^A, w) = E^{G_A}(S^p, \theta^A) + \gamma E^{G_{nNCF}}(S^p, w)$$
$$= \sum_{i=1}^{NP} (\gamma w^p(i, s_i^p) - log(p_G(x_i, \theta^A, s_i^p)))$$
$$+ \eta \sum_{i=1}^{NP} \sum_{j \in N_i} |s_i^p - s_i^p| (\gamma w^p(i, j) + \frac{e^{\beta \|x_i - x_j\|^2}}{D(i, j)}) \quad (27)$$

### F. Optimization

The optimization of the proposed NC-Cut is conducted iteratively, and estimates appearance models and NC jointly.

In step 1, the parameters of appearance models (GMMs) are estimated. For a given foreground ($s = 1$) or background ($s=0$) GMM, $FB(k, s) = \{x_i | k_i = k, s_i^p = s\}$ defines pixels for the $k$th GMM component; the mixture weights are $\pi(k, s) = |FB(k, s)| / \sum_{k=1}^{K} |FB(k, s)|$; the mean $\pi(k, s)$ is defined as the sample mean $\mu(k, s) = \sum_{x_i \in FB(k, s)} xx_i / |FB(k, s)|$; the covariance matrices $\sum(k, s)$ are estimated as the covariance of pixels' RGB values in $FB(k, s)$.

In step 2, the SRs set is updated by using the segmentation result ($S^p$) of the previous iteration: the regions with more than 80% pixels labeled as 0s will be added to SRs; in step 3, Algorithm 1 generates the NC and NCF based on image regions; in step 4, we apply the object and background regions to update

### Algorithm 2: Iterative NC cut

**Inputs:** image $x = \{x_i\}_{i=1}^{NP}$, and a user specified ROI
**Outputs:** $S^p = \{s_i^p \in \{0,1\}\}_{i=1}^{NP}$
**Initialization:**

$$s_i^p = \begin{cases} 1, if\ x_i\ is\ in\ the\ ROI \\ 0, otherwise \end{cases}, i = 1, \cdots, N^p$$

generate image regions $R = \{R_i\}_{i=1}^{N}$ using SLICO, compute the dissimilarity matrix ATs and indeterminacy matrix AIs
1: Learning the object and background GMMs' parameters $\theta^A$
2: Update the SRs
3: Compute background NC and NCF using Algorithm 1
4: NCF update
5: Update weights of links $w^p$ using Eqs. (23) and (24)
6: Apply max-flow algorithm to solve Eq. (27)
7: Repeat steps 1 to 6 until convergence the background NCF; in step 5, the pixel-wise weights of t-links and n-links are updated by using Eqs. (23) and (24); and the cost function is optimized by using the max-flow algorithm in step 6. All the steps will repeat until convergence (no change of the segmentation result).

**User editing.** If an interactive image segmentation approach cannot generate a satisfied result, further user editing is needed [1, 6]. For Algorithm 2, the user editing is to brush the wrongly labeled object and background superpixels and to update the GMMs and SRs; then the entire iterations (steps 1-7) of Algorithm 2 are applied. Notice that the image region set $R$, matrix ATs and AIs, and previous segmentation could be re-used.

## IV. EXPERIMENTAL RESULTS

### A. Dataset, metrics and parameter settings

The performance of the proposed NC-Cut method is validated using two datasets. The first dataset (DS1) is the widely used Grabcut [6] dataset which contains 50 images from the Berkley image dataset (BSD300) [18]. The second dataset (DS2) includes 215 images from MSRA [19] dataset. The manually marked regions of objects are used as the ground truth (GT).

The intersection-over-union (IoU) score [35], Error Rate (ERR), Rand Index (RI) [20], Global Consistency error (GCE) [21], and Boundary Displacement Error (BDE) [22] are commonly employed to assess the performance of the interactive segmentation methods. The IoU score for the object/background is defined as the ratio of the number of correctly labeled object/background pixels to the number of pixels labelled with the object/background in either the ground truth or the segmentation result. The average IoU over object and background are employed to evaluate the overall performance. The ERR counts the percentage of wrongly labeled pixels in ROI. The RI computes the fraction of pixel pairs having consistent labels in both the segmented result and the ground truth; which takes value in [0, 1], value 1 indicates that the seg-mentation result and the ground truth are exactly the same, and 0 indicates that they disagree on every pixel pairs. The GCE measures the consistency between the segmentation result and the ground truth; it takes value in [0, 1], and the value close to 0 indicates high accuracy. The BDE computes the average displacement of boundary pixels between the segmentation result and the ground truth; the displacement error of a boundary pixel is defined as the Euclidean distance between the pixel and its closest pixel on the boundary of the ground truth.

TABLE I
MAJOR SYMBOLS AND PARAMETERS.

| symbols/ parameters | description | value |
|---|---|---|
| $s^p$ | binary label for pixel | {0,1} |
| $N^p$ | number of pixels | / |
| $\theta^A$ | GMM parameters | / |
| K | number of GMM components | 5 |
| NCF | Neutro-Connectedness forest | / |
| $G_{nNCF}$ | graph built on the modified NCF | / |
| $w^p$ | weights of the links on $G_{nNCF}$ | / |
| $SR_s$ | set of region seeds | / |
| $(T,I,F)$ | the degree of truth, indeterminacy and falsity between two points | / |
| $h$ | inhomogeneity of region | [0, 1] |
| $\mu_T$ | strength of connectedness between two adjacent points | [0, 1] |
| $\mu_I$ | degree of indeterminacy for connectedness | [0,1] |
| $\preccurlyeq$ | a lexicographical order relation | / |
| $\gamma$ | degree of encouraging NC | [0, 1] |
| $p^{obj}$ | set of candidate object regions | / |
| $p^{bkg}$ | set of background regions | / |
| $\delta_t$ | stander deviation for computing $\mu_T$ | 30 |
| $\delta_B$ | stander deviation for selecting background seeds | 50 |
| $\delta_\gamma$ | stander deviation for computing $\gamma$ | 0.025 |
| $\varepsilon$ | threshold for selecting background seeds | 0.5 |

We compare the proposed NC-Cut algorithm with five state-of-the-art interactive segmentation algorithms: $MGC_{max}^{sum}$, Grabcut [6], MILCut [32], One-Cut [8] and pPBC [33]. $MGC_{max}^{sum}$ is a region seed-based approach implemented by following the $GC_{max}^{sum}$ method [5]; the weights of neighboring links (arcs) in $MGC_{max}^{sum}$ are calculated only based on local color difference (same as the proposed NC). For more information about link-weight estimation based on both local discontinuities and objects information, please refer [42]. The parameters of Grabcut [6], MILCut [32] and pPBC [33] are adopted from the related publications. In One-Cut [8], the number of color bins is set as $128^3$, the weight of smoothness is set as 5. Here, we would like to acknowledge the authors of MILCut, One-Cut and pPBC for providing the source code or executable program. The code of the Grabcut is available online (http://grabcut.weebly.com/code.html). All experiments are performed on windows-based PC equipped with a dual-core processor (2GHz) and 4GB memory.

The major symbols and parameters defined in this paper are summarized in Table 1. Constant $\delta_t$ controls the level of sensitivty of $\mu_T$ to the color difference between two regions, and is set to 30 by experiment; larger $\delta_t$ will make $\mu_T$ less sensitive to the differences. Constant $\delta_\gamma$ is the standard deviation used in the Gaussian function (Eq.(15)) to calculate the contribution ($\gamma$) of the NC-based term, and is chosen to be 0.025 by experiment. Parameters $\delta_B$ and $\varepsilon$ are used for background seeds selection (III(c)); and $\delta_B$ and $\varepsilon$ are set as 50 and 0.5 by experiment.

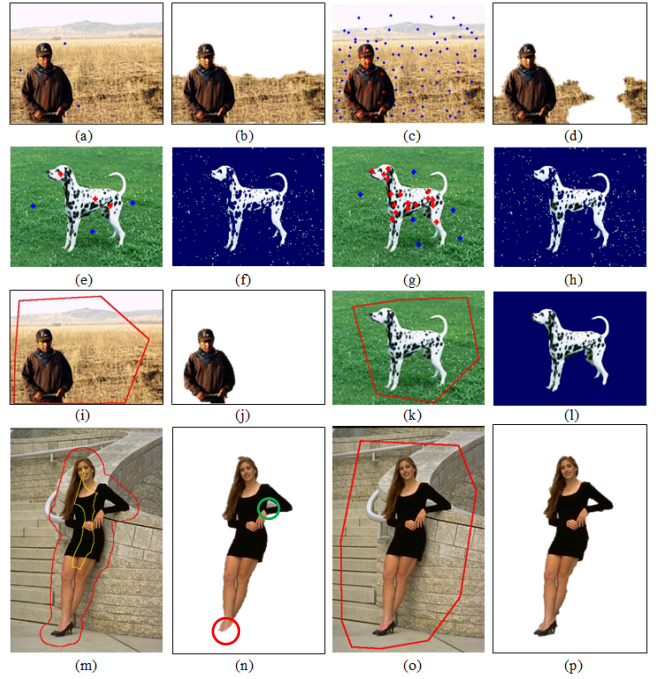

Fig. 8. (a) and (e) two sample images with 6 seeds each (3 object seeds red) and 3 background seeds blue)); (b) and (f) results of $MGC_{max}^{sum}$ on (a) and (e), respectively; (c) sample image with 92 seeds; (g) sample image with 23seeds; (d) and (h) results of $MGC_{max}^{sum}$ on (c) and (g), respectively; (i) and (k) two sample images with a ROI (red polygon) in each; (j) and (l) results of the proposed NC-Cut method on (i) and (j), respectively; (m) and (n) the seeds and results of $MGC_{max}^{sum}$; (o) and (p) the ROI and result of the proposed NC-Cut.

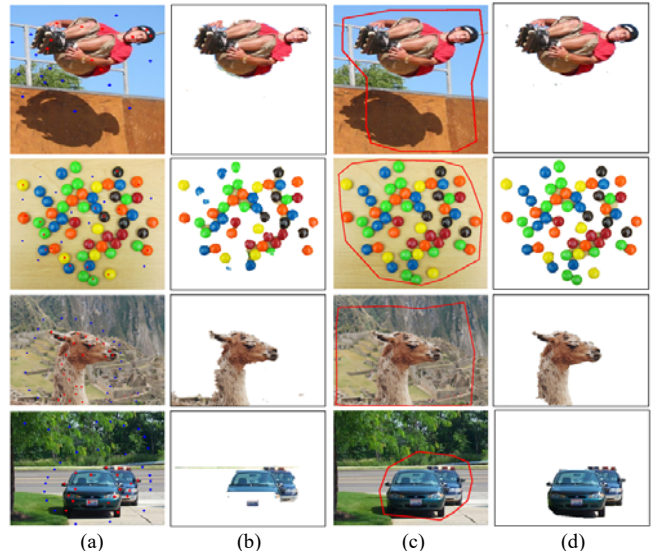

Fig. 9. (a) Four images with intensive user interactions (from top to down: 24, 32, 48 and 45 seeds); (b) results of $MGC_{max}^{sum}$ on (a); (c) four ROIs specified by user; (d) results of the proposed NC-Cut on (c).

TABLE II
SEGMENTATION PERFORMANCE OF $\text{MGC}_{\text{MAX}}^{\text{sum}}$ AND THE PROPOSED NC-CUT.

| Results | Methods | Interactions | Metrics | | | | Time (s) |
|---------|---------|--------------|---------|---|---|---|----------|
| | | | ERR(%) | RI | GCE | BDE | |
| Fig. 8(b) | $\text{GC}_{\text{max}}^{\text{sum}}$ | 6 seeds | 29.3 | 0.59 | 0.20 | 53.8 | 14.2 |
| Fig. 8(f) | $\text{GC}_{\text{max}}^{\text{sum}}$ | 6 seeds | 6.3 | 0.88 | 0.08 | 11.1 | 27.6 |
| Fig. 8(d) | $\text{GC}_{\text{max}}^{\text{sum}}$ | 92 seeds | 14.1 | 0.76 | 0.14 | 49.3 | 16.1 |
| Fig. 8(h) | $\text{GC}_{\text{max}}^{\text{sum}}$ | 23 seeds | 3.8 | 0.93 | 0.06 | 11.9 | 30.3 |
| Fig. 8(j) | NC-Cut | A polygon | 1.3 | 0.98 | 0.02 | 1.8 | 5.0 |
| Fig. 8(l) | NC-Cut | A polygon | 1.6 | 0.98 | 0.02 | 0.8 | 6.9 |
| Fig. 8(p) | NC-Cut | A polygon | 1.2 | 0.98 | 0.02 | 1.3 | 6.0 |

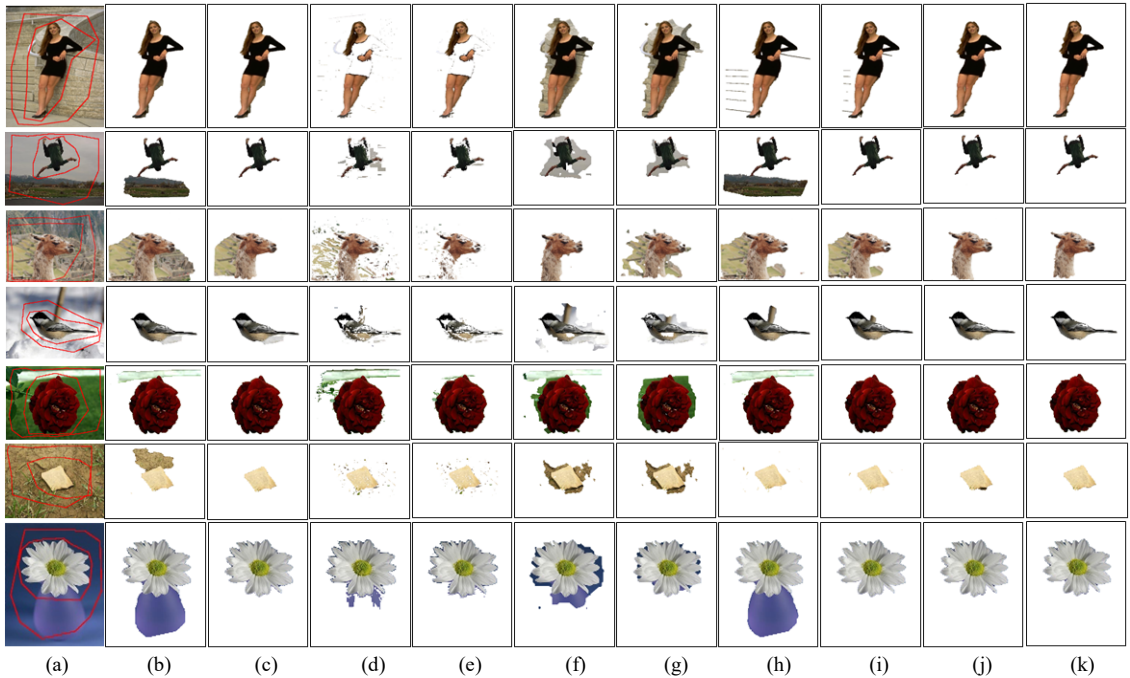

(a)    (b)    (c)    (d)    (e)    (f)    (g)    (h)    (i)    (j)    (k)

Fig. 10. (a) Seven sample images with two user-specified ROIs (tight and loose); (b) and (c) results of Grabcut using loose and tight ROIs, respectively; (d) and (e) results of One-Cut using loose and tight ROIs, respectively; (f) and (g) results of MILCut using loose and tight ROIs, respectively; (h) and (i) results of the pPBC using loose and tight ROIs, respectively; (j) and (k) results of the proposed NC-CUT using loose and tight ROIs, respectively.

### B. Interaction intensity

In this section, we will compare the proposed NC-Cut method with the $\text{MGC}_{\text{max}}^{\text{sum}}$ method. The $\text{MGC}_{\text{max}}^{\text{sum}}$ is a seed-based interactive segmentation method which employs the relative fuzzy connectedness (RFC) and the Graph cuts. Two seed generation methods were used in [5]: the first method needed user to specify both object and background seeds, and the second method obtained object and background seeds automatically by eroding and dilating the ground truth (only for evaluation), respectively. The qualitative and quantitative results of $\text{MGC}_{\text{max}}^{\text{sum}}$ with different interaction intensities are shown in Figs. 8 and 9, and Table II, respectively.

If only few seeds are specified in Figs. 8(a) and (e), the $\text{MGC}_{\text{max}}^{\text{sum}}$ method mislabeles many pixels (Figs. 8(b) and (f)); if more seeds are utilized, the results of $\text{MGC}_{\text{max}}^{\text{sum}}$ will be improved (Figs. 8(d) and (h)). As shown in Figs. 8(i) – (l), the proposed NC-Cut can achieve quite accurate result with only a loose ROI for each image.

Fig. 8 (m) shows the automatic seed generation results of $\text{MGC}_{\text{max}}^{\text{sum}}$ by using eroding and dilating operations on the ground truth, the pixels inside the yellow boundary are viewed as object seeds and outside the red boundary are set as background seeds; Fig. 8 (n) demonstrates that $\text{MGC}_{\text{max}}^{\text{sum}}$ cannot exclude the isolated background region in the green circle, and misclassified the foreground region in the red circle. The proposed NC-Cut can obtain more accurate result (Fig. 8(p)) than that of $\text{MGC}_{\text{max}}^{\text{sum}}$ with only a loose ROI because the proposed linking operation enforces object label consistency and pruning operation breaks the link between the isolated regions and their parents. More segmentation results of the $\text{GC}_{\text{max}}^{\text{sum}}$ and the proposed NC-Cut are shown in Fig. 9.

As shown in Fig. 9 and Table II, the $\text{MGC}_{\text{max}}^{\text{sum}}$ mehod needs intense user interactions to achieve a good performance, e.g., 92 seeds were specified in Fig. 8(c) to reduce the ERR to 14.1%. However, for Figs. 8(j), (l) and (p), the proposed NC-Cut has extremely low ERRs (**1.3%**, **1.6%** and **1.2** %), quite high RIs (**0.98**, **0.98** and **0.98**), and low GCEs (**0.02**, **0.02** and, **0.02**) and BDEs (**1.8**, **0.8** and **1.3**).

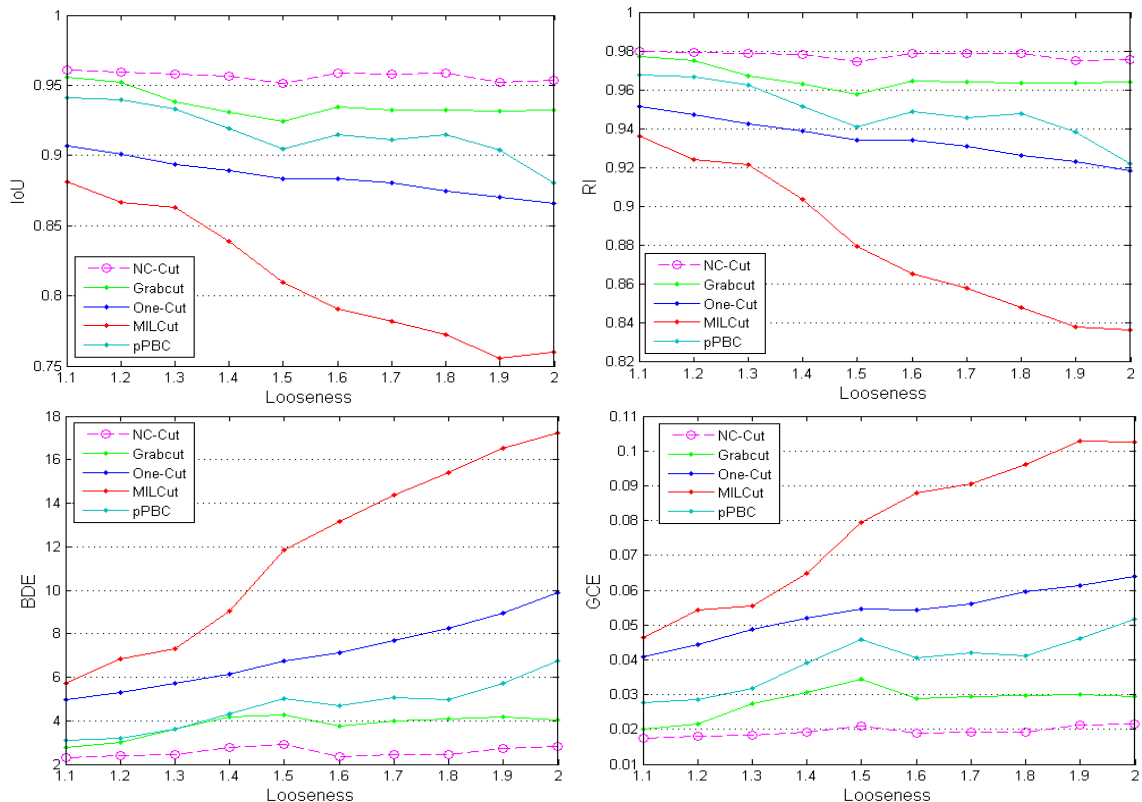

Fig. 11. Segmentation performance according to different ROI looseness.

### C. Interaction dependence

Now, we compare the proposed NC-Cut method with other four ROI-based interactive methods: Grabcut [6], One-Cut [8], MILCut [32] and pPBC [33].

Fig. 10 (a) shows seven natural images; each image has one loose ROI and one tight ROI specified by user. Fig. 10(b) demonstrates that when the ROIs are loose, the Grabcut method wrongly labels many pixels; Fig. 10(c) shows that when the ROIs are tight, the Grabcut produces quite accurate results. As shown in Figs. 10(d) and (e), One-Cut is less sensitive to the initial ROI than Grabcut, and can obtain better results than Grabcut on the last two images with loose ROIs; however, it cannot generate better results than Grabcut on the images with complicated object appearance (the third and fourth images). The MILCut method may not always produce better results using tight ROI than using loose ROI (third image), and Figs.10(f) and (g) demonstrate that MILCut is more sensitive to the initial ROI than Grabcut and One-Cut. In Figs. 10(h) and (i), the pPBC obtains better results than the other three methods, but it is still sensitive to the initial ROI; while the proposed NC-Cut method can produce accurate results (Figs. 10(j) and (k)) for all the images, and generates almost the same segmentation results using either tight or loose ROIs.

In order to evaluate the five methods' level of sensitivity to different ROIs quantitatively, we generate 10 groups of bounding boxes with different looseness automatically. We use the bounding box of the ground truth as the baseline, and set its looseness to 1; then move the four sides of the bounding box toward image borders to increase the looseness. The amount of move is proportional to the margin between the side and the image border. The looseness of a bounding box is defined as the ratio of area of the new bounding box to the area of the baseline bounding box. Twenty eight images having bounding box with looseness at least 2 are selected from the two datasets. As shown in Fig.14, the proposed NC-Cut is much less sensitive to the looseness of ROI than the other four methods.

### D. Overall performance

The proposed NC-Cut and other four state-of-the-art ROI-based methods are compared by utilizing the two datasets with predefined tight ROIs and loose ROIs. We also compare the proposed method without using the indeterminacy (NC-Cut$_0$) with the proposed NC-Cut to demonstrate the impacts of the indeterminacy on segmentation performance. We specify loose ROIs in the original images by drawing polygons to include more background pixels. As shown in Tables III and IV, using datasets DS1 and DS2, Grabcut obtains better average results with tight ROIs than those with loose ROIs; using dataset DS2, the average performances of One-Cut and MILCut drop dramatically when the ROIs change from tight to loose.

The ERRs with tight ROIs might be higher than those with loose ROIs because a good segmentation method will produce similar segmented results using both loose and tight ROIs; however, the sizes of tight ROIs are much smaller than the sizes of loose ROIs; i.e., ERR is not a good metric for comparing the performances for the same image with different sizes of ROIs. The proposed NC-Cut method outperforms Grabcut, One-Cut, MILCut and pPBC in all metrics (IoU, ERR, RI, GCE and BDE) using the two datasets with loose ROIs; and obtains similar

TABLE III
AVERAGE PERFORMANCE ON DATASET DS1

| Methods | Tight ROIs | | | | | Loose ROIs | | | | |
|---|---|---|---|---|---|---|---|---|---|---|
| | ERR(%) | RI | GCE | BDE | IoU | ERR(%) | RI | GCE | BDE | IoU |
| Grabcut | 12.8 | **0.94** | **0.04** | 7.7 | **0.91** | 13.9 | 0.90 | 0.06 | 12.2 | 0.85 |
| One-Cut | 23.5 | 0.88 | 0.06 | 10.4 | 0.80 | 15.8 | 0.88 | 0.08 | 12.7 | 0.80 |
| MILCut | 30.2 | 0.85 | 0.12 | 16.8 | 0.80 | 17.0 | 0.87 | 0.11 | 13.2 | 0.82 |
| pPBC | **12.1** | **0.94** | 0.05 | **6.8** | **0.91** | 9.8 | 0.92 | 0.06 | 8.8 | 0.83 |
| NC-Cut$_0$ | 18.2 | 0.91 | 0.05 | 14.1 | 0.85 | 11.3 | 0.91 | **0.05** | 12.1 | 0.85 |
| NC-Cut | 12.7 | **0.94** | 0.05 | 7.3 | **0.91** | **8.1** | **0.94** | **0.05** | **8.5** | **0.90** |

Bold represents the best result(s) in the same column; NC-Cut$_0$ is the proposed method without using the indeterminacy.

TABLE IV
AVERAGE PERFORMANCE ON DATASET DS2

| Methods | Tight ROIs | | | | | Loose ROIs | | | | |
|---|---|---|---|---|---|---|---|---|---|---|
| | ERR(%) | RI | GCE | BDE | IoU | ERR(%) | RI | GCE | BDE | IoU |
| Grabcut | **3.4** | **0.98** | **0.02** | 1.9 | **0.97** | 3.9 | 0.96 | 0.03 | 3.9 | 0.94 |
| One-Cut | 9.4 | 0.94 | 0.05 | 4.5 | 0.92 | 11.7 | 0.88 | 0.09 | 12.2 | 0.86 |
| MILCut | 19.8 | 0.89 | 0.08 | 8.5 | 0.86 | 31.9 | 0.73 | 0.17 | 24.7 | 0.67 |
| pPBC | 4.0 | **0.98** | **0.02** | 2.1 | **0.97** | 7.2 | 0.93 | 0.04 | 6.1 | 0.91 |
| NC-Cut$_0$ | 4.6 | 0.97 | **0.02** | 2.4 | 0.96 | 2.5 | 0.97 | **0.02** | **2.3** | **0.96** |
| NC-Cut | 3.5 | **0.98** | **0.02** | **1.8** | **0.97** | **2.3** | **0.98** | **0.02** | **2.3** | **0.96** |

Bold represents the best result(s) in the same column; NC-Cut$_0$ is the proposed method without using the indeterminacy.

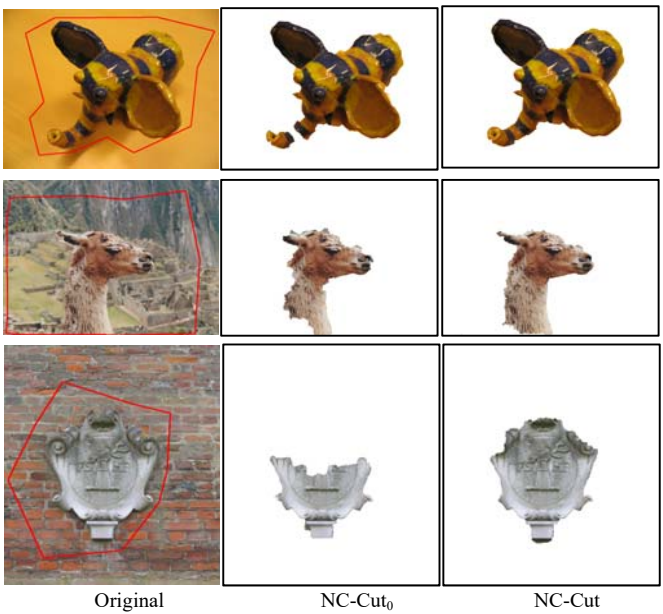

Original     NC-Cut$_0$     NC-Cut

Fig. 12. Results of NC-Cut$_0$ and NC-Cut.

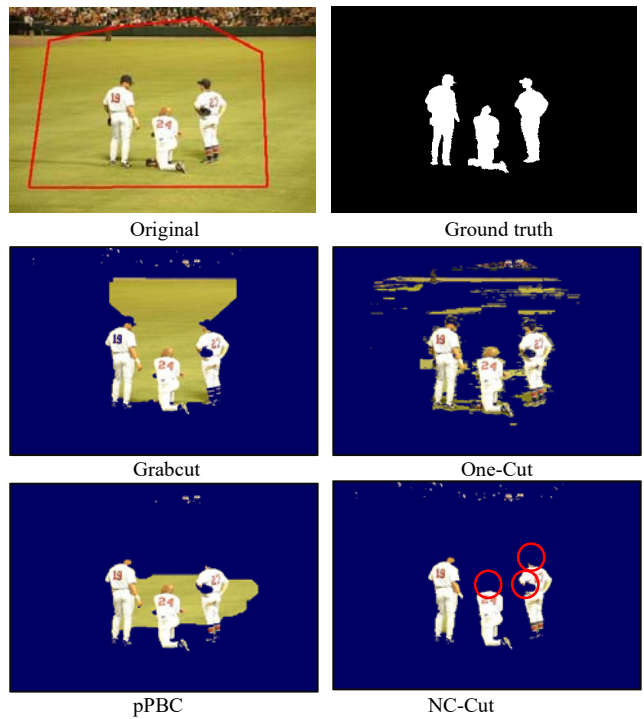

Original     Ground truth

Grabcut     One-Cut

pPBC     NC-Cut

Fig. 13. Failure case.

results with Grabcut and pPBC, and much better results than One-Cut and MILCut using tight ROIs. The average IoU, RI, GCE and BDE of the NC-Cut using the two datasets are almost the same with both the loose or tight ROIs, which implies that the proposed approach is not sensitive to the sizes of the initial ROIs; and this is the advantage which cannot be demonstrated by any existing state-of-the-art methods.

The only difference between NC-Cut$_0$ and NC-Cut is that the indeterminacy ($I$) is not utilized in NC-Cut$_0$ (i.e., $I = 0$). As shown in Table III and Fig. 12, NC-Cut$_0$ cannot achieve good performance as NC-Cut with both the loose or tight ROIs on DS1. Because most images in DS2 have homogenous background, the segmentation performance of NC-Cut$_0$ doesn't degrade much. Extensive experiment results of the proposed

NC-Cut, Grabcut, One-Cut, MILCut and pPBC methods are shown in Fig. 14. NC-Cut can avoid shrinking problem on many images because of the using of global properties; however, the contribution of the global term $E^{GnNCF}$ in NC-Cut is determined by the average indeterminacy of the connectedness (Eq. (15)). If an image has high level of average indeterminacy (small $\gamma$), NC-Cut will transfer more control to the appearance term ($E^{GA}$) which may result in the shrinking problem as Grabcut and Graph cuts on some images (Fig. 13). How to solve such problem will be studied in the future.

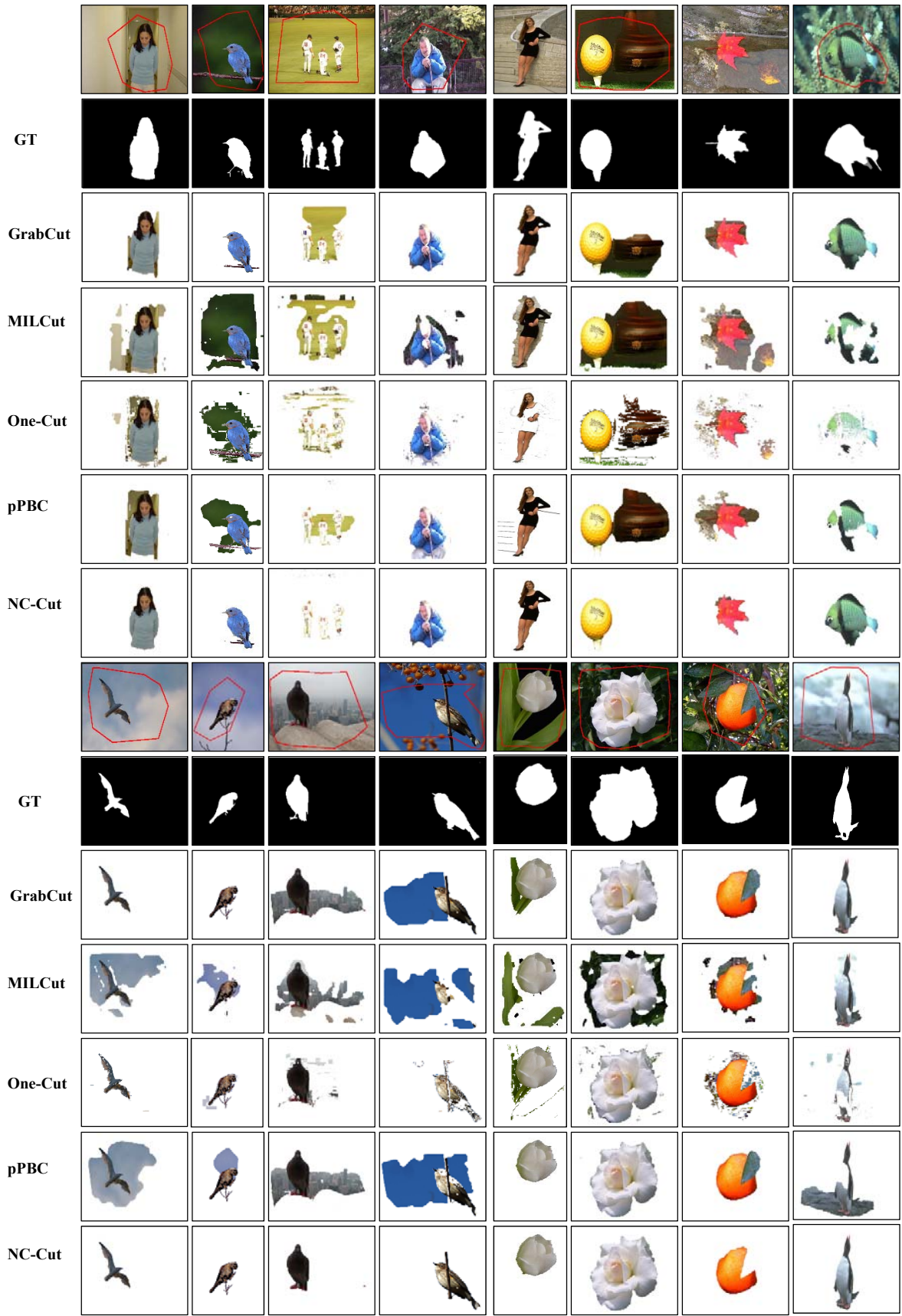

Fig. 14. Extensive segmentation results on natural images.

## V. CONCLUSION AND FUTURE WORK

In this paper, we generalize Neutro-Connectedness to model the topology of image regions, and propose a novel NC-based hybrid interactive segmentation method, NC-Cut, which needs much less user interaction than seed-based methods, and is less sensitive to the initial ROI than state-of-the-art ROI-based methods. There are three main contributions in this work: *first*, the generalized region-based NC is independent of the top-down priors of objects, and can model image topology with indeterminacy measurement; *second*, we update NCF based on the automatically detected object and background regions to enforce object label consistency and to exclude isolated background regions having low connectedness with the background seed regions; *third*, we formulate the proposed hybrid interactive segmentation approach, NC-Cut, using both pixel-wise appearance models and region-based NC, which can overcome the two problems of initial ROI-dependence and intense user interactions in interactive image segmentation approaches.

In the future, we will explore the possibility of applying the proposed Neutro-Connectedness to saliency detection, unsupervised image segmentation, object tracking, etc.

## REFERENCES

[1] Y. Boykov, and M. P. Jolly, "Interactive graph cuts for optimal boundary & region segmentation of objects in ND images," in Proc. IEEE ICCV, 2001, pp. 105-112.

[2] S. Vicente, V. Kolmogorov, and C. Rother, "Graph cut based image segmentation with connectivity priors," in Proc. IEEE CVPR, 2008, pp. 1-8.

[3] J. Stuhmer, P. Schroder, and D. Cremers, "Tree Shape Priors with Connectivity Constraints using Convex Relaxation on General Graphs," in Proc. IEEE ICCV, 2013, pp. 2336-2343.

[4] A. K. Sinop, and L. Grady, "A Seeded Image Segmentation Framework Unifying Graph Cuts And Random Walker Which Yields A New Algorithm," in Proc. IEEE ICCV, 2007, pp. 1-8.

[5] K. C. Ciesielski, P. A. Miranda, A. X. Falcão et al., "Joint graph cut and relative fuzzy connectedness image segmentation algorithm," Med. Image Anal., vol. 17, no. 8, pp. 1046-1057, 2013.

[6] C. Rother, V. Kolmogorov, and A. Blake, "Grabcut: Interactive foreground extraction using iterated graph cuts," ACM Trans. Graph., vol. 23, pp. 309-314, 2004.

[7] V. Lempitsky, P. Kohli, C. Rother et al., "Image segmentation with a bounding box prior," in Proc. IEEE ICCV, 2009, pp. 277-284.

[8] M. Tang, L. Gorelick, O. Veksler et al., "GrabCut in One Cut," in Proc. IEEE ICCV, 2013, pp. 1769-1776.

[9] M. Kass, A. Witkin, and D. Terzopoulos, "Snakes: Active contour models," Int. J. Comput. Vis., vol. 1, no. 4, pp. 321-331, 1988.

[10] R. Malladi, J. Sethian, and B. C. Vemuri, "Shape modeling with front propagation: A level set approach," IEEE Trans. Pattern Anal. Mach. Intell., vol. 17, no. 2, pp. 158-175, 1995.

[11] S. Osher, N. Paragios, Geometric level set methods in imaging, vision, and graphics, Springer Science & Business Media, 2003.

[12] A. X. Falcão, J. K. Udupa, S. Samarasekera et al., "User-steered image segmentation paradigms: Live wire and live lane," Graph. Models, vol. 60, no. 4, pp. 233-260, 1998.

[13] E. N. Mortensen, and W. A. Barrett, "Interactive segmentation with intelligent scissors," Graph. Models, vol. 60, no. 5, pp. 349-384, 1998.

[14] L. Chen, H. D. Cheng, and J. P. Zhang, "Fuzzy Subfiber and Its Application to Seismic Lithology Classification," Inf. Sci. Applications, vol. 1, no. 2, pp. 77-95, Mar, 1994.

[15] J. K. Udupa, and S. Samarasekera, "Fuzzy connectedness and object definition: Theory, algorithms, and applications in image segmentation," Graph. Models, vol. 58, no. 3, pp. 246-261, May, 1996.

[16] M. Xian, "A Fully Automatic Breast Ultrasound Image Segmentation Approach Based On Neutro-Connectedness," in ICPR, 2014, pp. 2495-2500.

[17] R. Achanta, A. Shaji, K. Smith et al., "SLIC superpixels compared to state-of-the-art superpixel methods," IEEE Trans. Pattern Anal. Mach. Intell., vol. 34, no. 11, pp. 2274-2282, 2012.

[18] D. Martin, C. Fowlkes, D. Tal, J. Malik, A database of human segmented natural images and its application to evaluating segmentation algorithms and measuring ecological statistics, in: IEEE ICCV, 2001, pp. 416-423.

[19] T. Liu, Z. Yuan, J. Sun et al., "Learning to detect a salient object," IEEE Trans. Pattern Anal. Mach. Intell., vol. 33, no. 2, pp. 353-367, 2011.

[20] W. M. Rand, "Objective criteria for the evaluation of clustering methods," J. Am. Stat. Assoc., vol. 66, no. 336, pp. 846-850, 1971.

[21] D. R. Martin, An Empirical Approach to Grouping and Segmentation: Computer Science Division, University of California, 2003.

[22] J. Freixenet, X. Muñoz, D. Raba et al., "Yet another survey on image segmentation: Region and boundary information integration," in Proc. IEEE ECCV, 2002, pp. 408-422.

[23] H.D. Cheng, and Y. Sun, "A hierarchical approach to color image segmentation using homogeneity," IEEE Trans. Image Process., vol. 9, no. 12, pp. 2071-2082, 2000.

[24] S. Xiang, F. Nie, and C. Zhang, "Interactive natural image segmentation via spline regression," IEEE Trans Image Process, vol. 18, no. 7, pp. 1623-32, Jul, 2009.

[25] T. V. Spina, P. A. de Miranda, and A. X. Falcao, "Hybrid approaches for interactive image segmentation using the live markers paradigm," IEEE Trans Image Process, vol. 23, no. 12, pp. 5756-69, Dec, 2014.

[26] A. X. Falcao, J. K. Udupa, and F. K. Miyazawa, "An ultra-fast user-steered image segmentation paradigm: live wire on the fly," IEEE Trans Med Imaging, vol. 19, no. 1, pp. 55-62, Jan, 2000.

[27] P. A. Miranda, A. X. Falcao, and T. V. Spina, "Riverbed: a novel user-steered image segmentation method based on optimum boundary tracking," IEEE Trans Image Process, vol. 21, no. 6, pp. 3042-52, Jun, 2012.

[28] A. X. Falcao, J. Stolfi, and R. de Alencar Lotufo, "The image foresting transform: theory, algorithms, and applications," IEEE Trans Pattern Anal Mach Intell, vol. 26, no. 1, pp. 19-29, Jan, 2004.

[29] S. Han, W. Tao, D. Wang et al., "Image segmentation based on GrabCut framework integrating multiscale nonlinear structure tensor," IEEE Trans Image Process, vol. 18, no. 10, pp. 2289-302, Oct, 2009.

[30] P. Scheunders, "A multivalued image wavelet representation based on multiscale fundamental forms," IEEE Trans Image Process, vol. 11, no. 5, pp. 568-75, 2002.

[31] A. Protiere, and G. Sapiro, "Interactive image segmentation via adaptive weighted distances," IEEE Trans Image Process, vol. 16, no. 4, pp. 1046-57, Apr, 2007.

[32] J. Wu, Y. Zhao, J.-Y. Zhu, S. Luo, Z. Tu, Milcut: A sweeping line multiple instance learning paradigm for interactive image segmentation, in: IEEE CVPR, 2014, pp. 256-263.

[33] M. Tang, I.B. Ayed, Y. Boykov, Pseudo-Bound Optimization for Binary Energies, in: ECCV, Springer, 2014, pp. 691-707.

[34] S.D. Jain, K. Grauman, "Predicting Sufficient Annotation Strength for Interactive Foreground Segmentation", in IEEE ICCV, 2013, pp. 1313-1320.

[35] Y. Zeng, D. Samaras, W. Chen et al., "Topology cuts: A novel min-cut/max-flow algorithm for topology preserving segmentation in N–D images," Computer Vision and Image Understanding, vol. 112, no. 1, pp. 81-90, 2008.

[36] S. Nowozin, and C. H. Lampert, "Global interactions in random field models: A potential function ensuring connectedness," SIAM Journal on Imaging Sciences, vol. 3, no. 4, pp. 1048-1074, 2010.

[37] V. Gulshan, C. Rother, A. Criminisi et al., "Geodesic star convexity for interactive image segmentation." In IEEE CVPR, 2010, pp. 3129-3136.

[38] Veksler, Olga. "Star shape prior for graph-cut image segmentation."Computer Vision–ECCV 2008. Springer Berlin Heidelberg, 2008. 454-467.

[39] B. L. Price, B. Morse, and S. Cohen, "Geodesic graph cut for interactive image segmentation." In IEEE CVPR, 2010, pp. 3161-3168.

[40] K. C. Ciesielski, J. K. Udupa, A. X. Falcao et al., "Fuzzy Connectedness Image Segmentation in Graph Cut Formulation: A Linear-Time Algorithm and a Comparative Analysis," Journal of Mathematical Imaging and Vision, vol. 44, no. 3, pp. 375-398, Nov, 2012.

[41] P. E. Rauber, A. X. Falcao, T. V. Spina et al., "Interactive segmentation by image foresting transform on superpixel graphs," in: Conference on Graphics, Patterns and Images (SIBGRAPI), 2013, pp. 131-138.

[42] P. A. V. de Miranda, A. X. Falcão, and J. K. Udupa, "Synergistic arc-weight estimation for interactive image segmentation using graphs,"

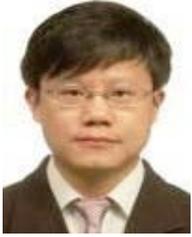

**Min Xian** received the B.S. degree in information security and M.S. degree in pattern recognition and intelligence system from Harbin Institute of Technology, China, 2008 and 2011, respectively. He is currently a Ph.D. candidate in computer science at Utah State University. His research interest includes image segmentation, biomedical image processing, pattern recognition, and the theory, algorithm and applications of Neutro-Connectedness.

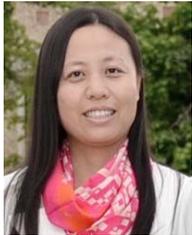

**Yingtao Zhang** received M.S. degree from Computer Science School of Harbin Institute of Technology, Harbin, China, 2004, and Ph.D. degree in Pattern Recognition and Intelligence System from Harbin Institute of Technology, Harbin, China, 2010. Now, she is an associate professor, School of Computer Science and Technology, Harbin Institute of Technology, Harbin, China. Her research interests include pattern recognition, computer vision, fuzzy logic and neutrosophic logic, and medical image processing.

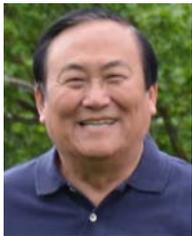

**Heng-Da Cheng** received the Ph.D. degree in Electrical Engineering from Purdue University (Supervisor: Prof. K.S. Fu), West Lafayette, IN, 1985. Now, he is a Full Professor, Department of Computer Science, and an Adjunct Full Professor, Department of Electrical Engineering, Utah State University, Logan, Utah. Dr. Cheng is an Adjunct Professor and Doctorial Supervisor of Harbin Institute of Technology. He is also a Guest Professor of the Institute of Remote Sensing Application, Chinese Academy of Sciences, a Guest Professor of Wuhan University, a Guest Professor of Shantou University, and a Visiting Professor of Northern Jiaotong University.

Dr. Cheng has published more than 350 technical papers and is the Co-editor of the book, Pattern Recognition: Algorithms, Architectures and Applications (World Scientific Publishing Co., 1991).

His research interests include image processing, pattern recognition, computer vision, artificial intelligence, medical information processing, fuzzy logic, genetic algorithms, neural networks, parallel processing, parallel algorithms, and VLSI architectures.

Dr. Cheng was the General Chair of the 11th Joint Conference on Information Sciences (JCIS 2008), the General Chair of the 10th Joint Conference on Information Sciences (JCIS 2007), the General Chair of the Ninth Joint Conference on Information Sciences (JCIS 2006), the General Chair of the Eighth Joint Conference on Information Sciences (JCIS 2005), etc. He served as program committee member and Session Chair for many conferences, and as reviewer for many scientific journals and conferences. Dr. Cheng has been listed in Who's Who in the World, Who's Who in America, Who's Who in Communications and Media, etc.

Dr. Cheng is also an Associate Editor of Pattern Recognition, an Associate Editor of Information Sciences and Associate Editor of New Mathematics and Natural Computation.

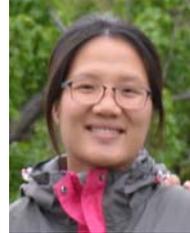

**Fei Xu** received B.S. degree in Computer Science and Technology from Northeast Normal University, Changchun, China, 2009, and the M.S. degree in Pattern Recognition and Intelligence System from Harbin Institute of Technology, Harbin, China, 2011. She is now a Ph.D. student in the Department of Computer Science, Utah State University. Her research interests include Saliency detection, pattern recognition, machine learning, and neutrosophic logic.

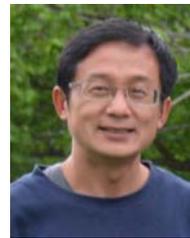

**Jianrui Ding** received B.S. degree in Computer Science from Harbin Institute of Technology, Harbin, China, 1995, M.S. degree and Ph.D. degree in Pattern Recognition and Intelligence System from Harbin Institute of Technology, Harbin, China, 1998 and 2013, respectively. He is now an associate professor in Harbin Institute of Technology (Weihai), China. His research interests include medical image processing, pattern recognition, machine learning, and neutrosophic logic.